\documentclass[letterpaper, 10 pt, conference]{ieeeconf} 
\IEEEoverridecommandlockouts
\usepackage{cite}
\usepackage{amsmath,amssymb,amsfonts}
\usepackage{algorithmic}
\usepackage{subcaption}
\usepackage{graphicx}
\usepackage{textcomp}
\usepackage{xcolor}
\usepackage{booktabs}
\usepackage{gensymb}
\usepackage{multirow}
\usepackage{balance}

\def\BibTeX{{\rm B\kern-.05em{\sc i\kern-.025em b}\kern-.08em
    T\kern-.1667em\lower.7ex\hbox{E}\kern-.125emX}}

\usepackage{etoolbox}
\usepackage{caption}
\newcommand{\insertfig}{
\vspace{0.7cm}
\includegraphics[width=\linewidth]{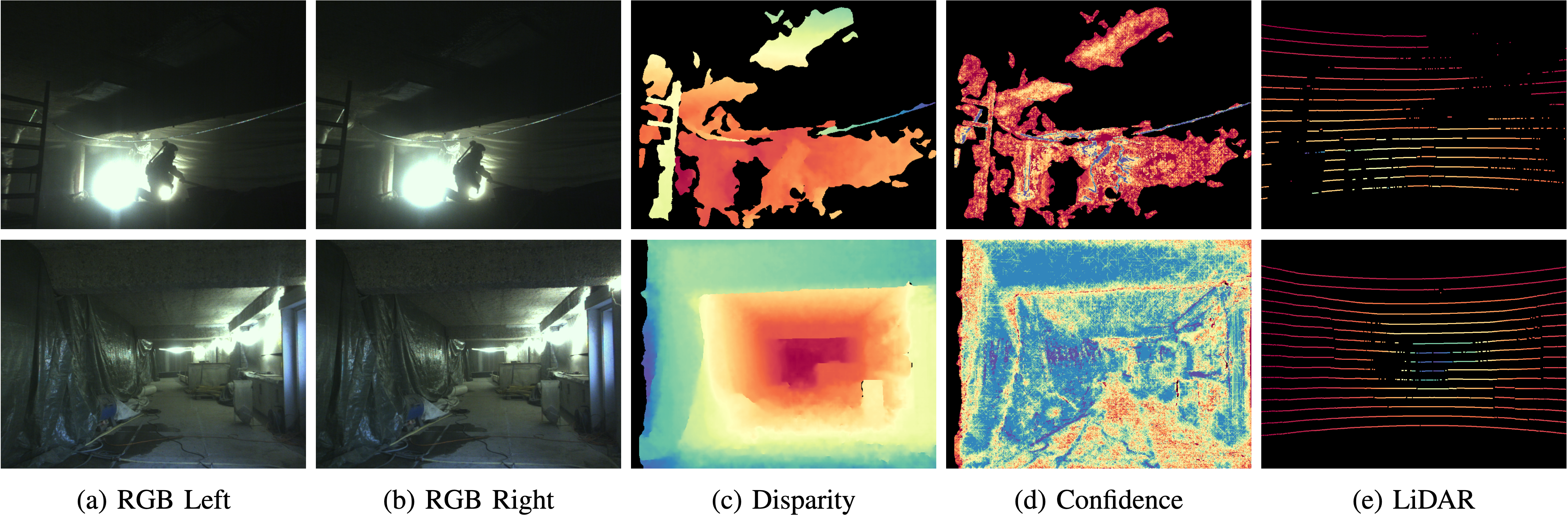}
\captionof{figure}{
\textbf{Data samples from our \emph{ShotcreteDepth} dataset.} RGB images originate from Roboception rc\_visard 160c camera, the disparity and confidence is computed internally by the camera, and point cloud from the Velodyne LiDAR projected to the image plane. Red color denotes the smallest and blue the largest values.\\~}
\label{fig:dataset-examples}
}

\makeatletter
\apptocmd{\@maketitle}{\centering\insertfig}{}{}%
\makeatother

\begin{document}
\title{\LARGE \bf
ShotcreteDepth: A Bi-modal Dataset for Robust Robotic Depth Perception in Shotcrete Construction Environments \\
\thanks{\textsuperscript{1}Technical University of Denmark, Department of Electrical and 
Photonics Engineering. \textsuperscript{2}Pioneer Centre for AI, Copenhagen, Denmark. 
\textsuperscript{3}Christiansen \& Essenbæk A/S. \textsuperscript{*}Corresponding author jagre@dtu.dk.
This work has been funded and supported by the EU Horizon Europe project “RoBétArmé” under the Grant 
Agreement 101058731. We express our gratitude to Søren Beyer Nielsen for designing and 3D printing the
dust-proof sensor housing.}
}

\author{
    Jakub Gregorek\textsuperscript{1,2*}, 
    Lars Arnold Dethlefsen\textsuperscript{1,2},
    Patrick Schmidt\textsuperscript{1,2},\\
    Mads Essenbæk\textsuperscript{3},
    Jonas Flink Bentzen\textsuperscript{3} and
    Lazaros Nalpantidis\textsuperscript{1,2}
}

\maketitle
\thispagestyle{empty}
\pagestyle{empty}
\begin{abstract}
We introduce \emph{ShotcreteDepth}, a bi-modal dataset from the construction domain 
that captures both an active shotcreting process and general construction 
environments. The dataset comprises stereo RGB imagery and LiDAR point 
clouds acquired under harsh real-world conditions, including high turbidity 
and poor illumination. Such conditions adversely affect sensor measurements, 
leading to incomplete and noisy observations that pose significant challenges 
for perception systems in autonomous applications. Alongside the dataset, 
we release a lightweight annotation tool designed for time-efficient labeling of 
LiDAR point clouds. ShotcreteDepth consists of 11,252 temporally synchronized 
data samples, of which 220 are annotated for evaluation purposes. The dataset 
supports research in stereo matching, depth completion, and depth estimation 
under conditions that closely reflect the operational complexities found in 
industrial settings. Project repository: https://github.com/dtu-pas/shotcrete-depth
\end{abstract}

\section{Introduction}
We are publishing a bi-modal dataset---\emph{ShotcreteDepth}---capturing a shotcreting environment,
shown in Fig.~\ref{fig:dataset-examples}, aimed at 
development and evaluation of stereo matching, depth completion and depth estimation methods. 
Furthermore, we are releasing a light-weight annotation tool for LiDAR point clouds. 
The application of sprayed concrete, ``shotcrete'', is a construction method which can be used for 
reinforcing unstable structures in underground mining or tunnel constructions, building complex geometries
or repairing damaged structures. The technology is mechanized to the large extend, yet the shotcrete
application is still laborious work performed manually. The motivation to automate this process is
rooted in achieving more consistent quality, ensuring compliance with the design requirements,
reducing material and water waste by eliminating over-spraying, and limiting exposure
of human workers to harmful quick-setting agents~\cite{robetarme}. The full automation
of the shotcreting process has many aspects~\cite{robetarme-segmentation,robetarme-segmentation-symposium,
robetarme-virtual-reality-learning,robetarme-shotcrete-simulation,robetarme-particle-modeling,
robetarme-path-planning,robetarme-profile-granding,robetarme-arepo}, one of which is advanced
perception for autonomous systems, which are to perform the application of shotcrete.
This work is aimed at the depth perception, which is relevant for navigation, mapping, 
obstacle avoidance, safety, measuring material deposition, compliance and process monitoring.
The shotcreting environment, characterized by turbidity, poses a challenge for 
robotic perception systems. The rebound phenomenon produces significant amount of 
airborne shotcrete dust concentrating in the enclosed environments.
The dust limits the visibility, negatively impacting camera based depth sensing, 
and causes laser scattering, challenging active depth sensors like LiDARs.
Development of reliable perception systems for the construction site domain, even more so 
for niche applications like shotcrete, suffer from lack of data.
The contributions of this paper are twofold and aiming to narrow this gap: 
\begin{itemize}
    \item We are releasing a shotcreting dataset capturing a niche construction environment 
    with the emphasis on depth perception. We are using this dataset to test various stereo, depth completion, and depth estimation baselines to elucidate its characteristics.
    \item We are providing a light-weight annotation tool for 3D point clouds allowing users
    to label unreliable depth measurements.
\end{itemize}

\section{Related Work}
\subsection{Relevant Datasets}
Only a few construction-related datasets contain any form of depth data, but they target a variety of tasks: SLAM~\cite{conslam},
segmentation~\cite{bridge-3d-segmentation,DING2024107964,buildingworld,pc-urban}, 
place recognition~\cite{conpr}, pose estimation~\cite{wfc-pose-estimation}, 
3D reconstruction~\cite{buildingworld}, object detection~\cite{DING2024107964}, 
depth estimation~\cite{DING2024107964}. 
When it comes to shotcreting 
or tasks related to the shotcreting process, existing datasets aim at segmentation~\cite{con-reb-seg}, 
structural performance~\cite{SJOLANDER2025111684} and deformations~\cite{cui2024deformation}.
Similar harsh environmental conditions to a certain extent, including turbidity, haziness, 
excessive or insufficient illumination can be observed in~\cite{seeing-through-fog,
fire-rescue-radar,dense-haze}.
To our knowledge, there is no existing dataset capturing the niche shotcreting domain
allowing development and evaluation of depth perception methods such as depth estimation,
completion and stereo matching.

\subsection{Depth Estimation}
Depth estimation is the task of estimating depth based on a provided monocular image.
Considering the data scarcity in this niche field, our interest is mainly on the methods that,
have been evaluated in the zero-shot context. Ranftl et al.~\cite{midas} pioneered the zero-shot
monocular depth estimation. Many of the recent depth estimation models are based on Vision Transformer (ViT) 
backbones~\cite{unidepth,unidepth2,depth-anything,depth-anything-v2,depth-pro,moge,moge2,
depth-anything-v3} initialized from DINOv2~\cite{dinov2}, and pre-trained ConvNext and 
ResNet backbones~\cite{metric3d,midas}. Others methods take the generative 
approach~\cite{marigold,marigold-e2e,better-depth,pixel-perfect,depth-fm,lotus,better-depth}. 
The depth estimation models may predict disparity~\cite{midas,lotus,depth-anything,depth-anything-v2}, 
relative depth estimates~\cite{marigold,depth-fm,depth-anything-v3,better-depth,pixel-perfect,moge}, 
metric depth~\cite{unidepth,unidepth2,metric3d,depth-pro,zoe-depth,moge2},
or can be fine-tuned for metric depth~\cite{depth-anything,depth-anything-v2,depth-anything-v3}. 
Whereas real-world labeled datasets are usually required to train the depth estimation 
models~\cite{midas,metric3d,unidepth,unidepth2}, some methods can take advantage of 
unlabeled data~\cite{depth-anything,depth-anything-v2,depth-anything-v3} and recently 
synthetic datasets have also started playing an important role~\cite{marigold,unidepth2,depth-pro,depth-fm,
lotus,better-depth,pixel-perfect,moge2,moge}.

\subsection{Depth Completion}
In the context of this paper we define depth completion to be the task of densifying sparse
depth, guided by a monocular image.
Zero-shot capable depth completion methods may benefit from the pre-trained priors of the
depth estimators~\cite{marigold-dc,capa,steered-marigold,needforspeed,Hyoseok_Kim_Byung-Ki_Oh_2025,
test-prompt-dc,lora-decoder-adaption,prompt-da} or stereo matching architectures~\cite{vpp4dc}.
Some of the recent state-of-the-art methods formulate the problem as test-time optimization.
Ke et al.~\cite{capa} applies low-rank adaptation and visual prompt tuning. Prompt tuning
was also explored by Jeong et al.~\cite{test-prompt-dc} and low-rank adaptation by Seo et 
al.~\cite{lora-decoder-adaption}. Diffusion-based methods~\cite{marigold-dc,steered-marigold,
Hyoseok_Kim_Byung-Ki_Oh_2025} use the sparse depth to iteratively guide the diffusion process.
Faster inference compared to the test-time optimization can be achieved by fine-tuning the 
diffusion model~\cite{needforspeed}. In contrast, Bartolomei et al.~\cite{vpp4dc} utilizes virtual pattern 
projector and re-trains stereo matching network~\cite{raft-stereo} achieving strong zero-shot
generalization. Liang et al.~\cite{dmd3c} proposes a distillation framework leveraging supervision
of the strong depth estimators, whereas Lin et al.~\cite{prompt-da} performs super-resolution targeting 
noisy low-resolution sensors. Finally, Zuo and Deng~\cite{ogni-dc} proposes an approach based on gated 
recurrent units iteratively refining depth gradients, depth integration and SPN enhancement. 
That work was further extended~\cite{omni-dc} introducing multi-resolution depth integrator
and Laplacian loss.

\subsection{Stereo Matching}
Stereo matching is the task of finding correspondences between images captured by a stereo camera
allowing to estimate depth of the observed scene. While Semi-Global Matching by 
Hirschmüller~\cite{sgm} became an industrial standard due to its favorable accuracy-efficiency 
trade-off~\cite{Nalpantidis2008_IJO}, the deep learning methods have dominated the task. Some of the methods aggregate 
cost volume~\cite{gcnet,cfnet,pcw-net,aanet}, which is often memory expensive. Other methods
perform recurrent disparity refinement~\cite{raft-stereo,any-stereo,mocha-stereo,igev-stereo}, 
which is less time-efficient. An alternative approach avoiding the construction of a cost-volume
is utilizing attention mechanisms establishing correspondences between stereo 
images~\cite{croco2-stereo}. The most recent methods combine the mentioned 
approaches~\cite{stereo-anywhere,defom-stereo,foundation-stereo} and simultaneously 
take advantage of the monocular foundational depth models~\cite{dinov2,depth-anything-v2}.

\section{Dataset}
\begin{table}[tb]
\caption{\textbf{\emph{ShotcreteDepth} dataset parameters.}
}
\label{tab:dataset-parameters}
\centering
\begin{tabular}{l|r}
\toprule
Parameter                  & Description/Value \\
\midrule
Stereo Camera              & Roboception rc\_visard 160 color \\
-- Field of View           & Horizontal: $61\degree$, Vertical: $48\degree$ \\
-- Image Resolution        & $1280 \times 960$ \\
-- Disparity Resolution    & $640 \times 480$ \\
\midrule
LiDAR                      & Velodyne PUCK              \\
-- Field of View           & Vertical: $30\degree$ \\
-- Depth Resolution        & $1280 \times 960$ \\
\midrule
Total Dataset Size         & 11252                   \\
Annotated Dataset Size     & 220                     \\
\bottomrule
\end{tabular}
\end{table}

\subsection{Overview}
Our dataset captures the environment where the shotcreting is performed, 
including scenes before, during and after shotcreting, as well as the general 
construction environment. The dataset is bi-modal, aiming to provide data
for development and evaluation of depth perception systems in conditions 
with challenging environmental conditions, including high turbidity and 
far-from-optimal illumination. It contains 11,252 temporally synchronized data samples, 
of which 220 are annotated for evaluation purposes. Selection of the evaluation set 
maximizes diversity, skipping similar consecutive frames. An overview of the dataset's 
parameters are presented in Tab.~\ref{tab:dataset-parameters}.

\subsection{Environment}
The construction site at which the dataset was collected exemplifies the type of 
environment autonomous systems will need to operate in, should this demanding task 
be automated. The workspace was sealed off from the outside, so illumination was provided 
exclusively by artificial lighting, and shotcrete particles accumulated rapidly in the air. 
The construction environments are often confined, as shotcreting is commonly done in 
tunnels or mines. These factors together create particularly harsh conditions for both 
human workers and equipment. To protect the sensors from shotcreting particles, 
they were mounted inside a custom 3D‑printed dust‑proof sensor housing, shown in Fig.~\ref{fig:hardware-setup}.

\subsection{Sensors \& Modalities}
\begin{figure}[t]
    \centering
    \includegraphics[width=0.5\linewidth]{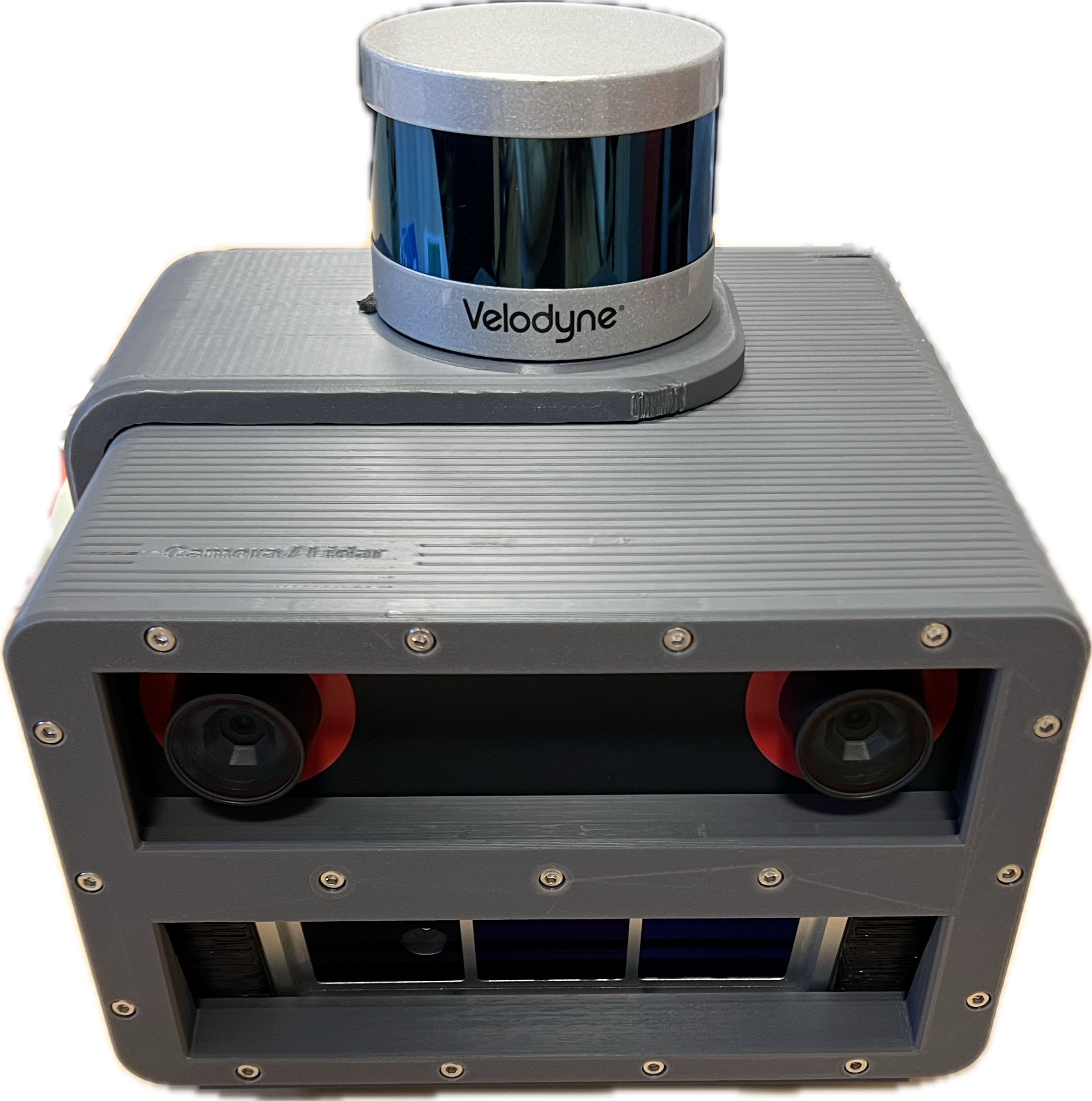}
    \caption{Our 3D printed dust-proof sensor housing containing the Roboception 
    rc\textunderscore visard 160 stereo camera and the Velodyne PUCK LiDAR.}
    \label{fig:hardware-setup}
\end{figure}

Our bi-modal dataset was captured using two sensors: Roboception 
rc\_visard 160c stereo camera with 4mm lens and Velodyne PUCK LiDAR. The stereo camera provides RGB 
images of resolution $1280\times960$ pixels, disparity and confidence maps of
resolution $640\times480$, which are computed by proprietary implementation of
Semi-Global Matching~\cite{sgm} running directly on the Nvidia Tegra K1 embedded 
in the camera. The Velodyne PUCK LiDAR measures depth up to 100m with typical
accuracy up to $\pm3cm$. The LiDAR has 16 scan lines, vertical field of view of 30°,
and operates on 903nm wavelength. The LiDAR was mounted approximately centered above
the stereo camera in our dust resistant housing. We have also performed tests with 
a solid-state LiDAR Neuvition Titan M1 (visible at the bottom of Fig.~\ref{fig:hardware-setup}, which did not seem to be well suited for 
this task. It struggled in the turbid environment to the extent of being unable 
to return any depth whatsoever. It could be explained by the longer wave-length
(1550nm) it operates with or internal processing not being tuned for these conditions.

\subsection{Calibration \& Synchronization}
The intrinsic camera parameters were estimated using the Roboception's calibration tool
which is built in the camera. The extrinsic parameters of the camera and LiDAR were 
obtained using Matlab's Camera LiDAR and Camera Calibration. The accurate LiDAR time-stamping 
was ensured by the PPS signal and NMEA messages received from a GPS module. The camera's 
clock was synchronized using PTP with a system clock of our acquisition computer (also synchronized
with GPS time). Our sensor setup acquires the modalities at different 
frequencies: RGB images at 25Hz, stereo disparity at 3Hz and LiDAR point clouds at 10 Hz. 
The RGB images and disparity from the stereo camera were matched with the temporally 
closest point cloud producing only complete sets containing all modalities
(cf. Fig.~\ref{fig:dataset-examples}).

\subsection{Influence of Turbidity}

Visibility worsens in turbid environments, which typically makes stereo
matching more difficult. Nevertheless, LiDAR seems to be affected more severely and
the shotcrete particles dispersed in the air seem to degrade the quality of depth measurements
in situations with reasonable visibility. The impact on LiDAR is three-fold:
\begin{enumerate}
    \item \textit{Missing data.} The large amount of shotcrete particles in the air may ``blind"
    the LiDAR either partially or even entirely.
    \item \textit{Noisy measurements.} Point clouds become visually more noisy during shotcreting.
    \item \textit{Dust clouds.} Shotcrete particles become visible in the LiDAR point clouds,
    while being mostly translucent for the camera and human eyes.
\end{enumerate}

\subsection{Point Cloud Annotation Tool}
\begin{figure}[t]
    \centering
    \includegraphics[width=0.8\linewidth]{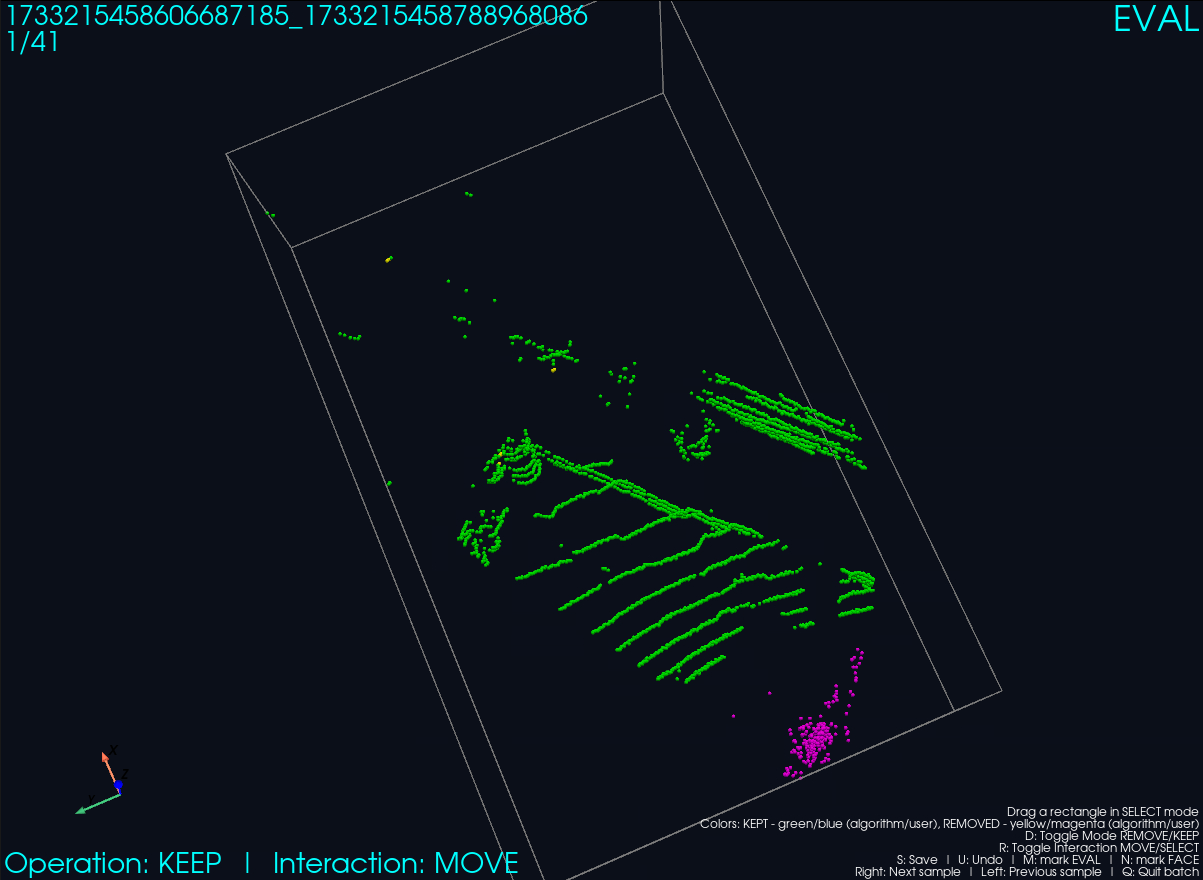}\\
    \smallskip
    \includegraphics[width=0.8\linewidth]{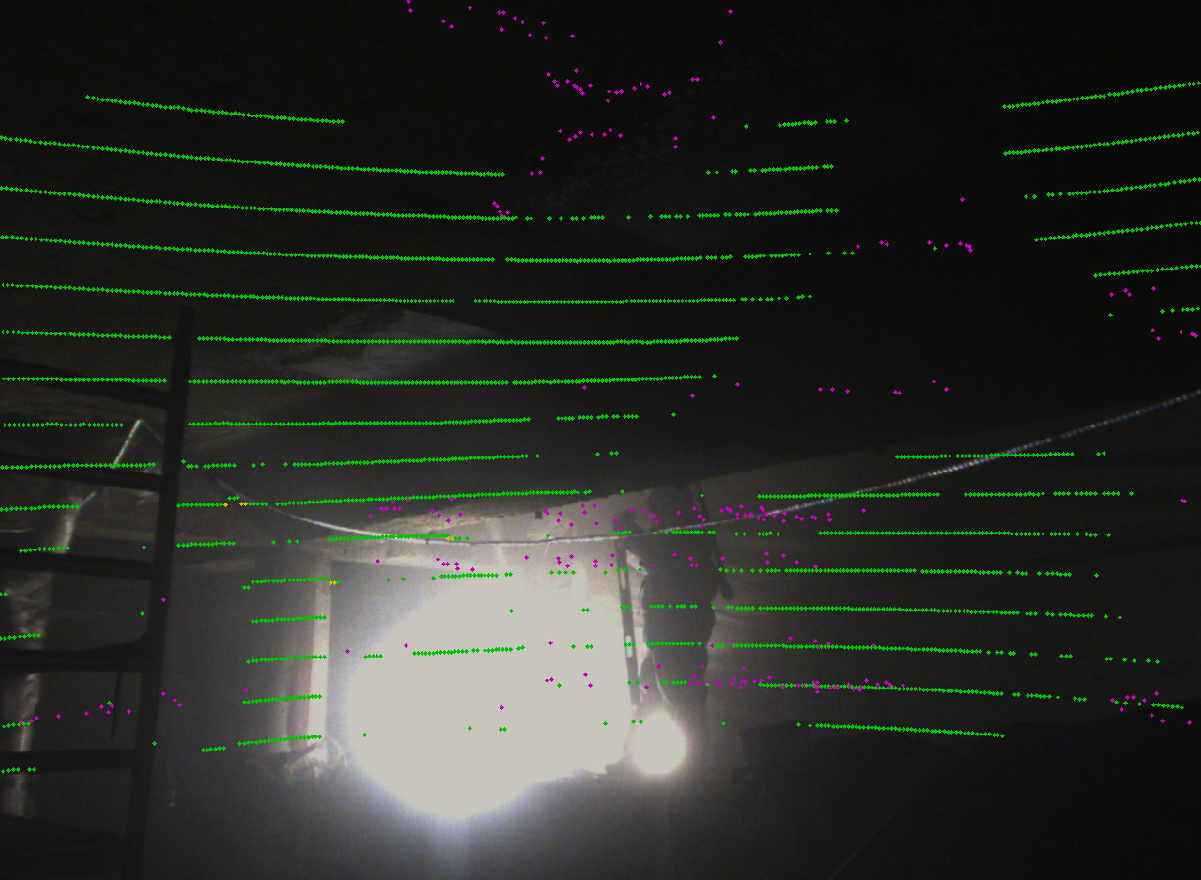}
	\caption{\textbf{The Annotation Tool} we are releasing with the dataset. Upper image:
    3D view of the LiDAR point cloud. Lower image: point cloud overlaid on top of the left
    camera image. The green points are ``kept" while the purple color denotes the dust 
    cloud which is to be excluded from evaluation data.}
    \label{fig:annotation-tool}
\end{figure}

We developed an annotation tool, allowing us to manually annotate the dust clouds of shotcrete
observed by the LiDAR. The tool allows the user to assign each point in the point cloud a 
``kept by user" or ``removed by user" label. Additionally we define the labels
``kept by algorithm" and ``removed by algorithm", which are to be used for annotations performed
programmatically. The user interface consists of two views (depicted in 
Fig.~\ref{fig:annotation-tool}): the first view displays the LiDAR point cloud in 3D, whereas the second view displays 
the projection of the point cloud on top of the left RGB picture. The user can switch modes 
using shortcuts to rotate or annotate the displayed point cloud following the selected 
operation, keeping or removing the points. Furthermore, the tool allows the user to 
flag the image for evaluation. We annotated only the point clouds of the dataset samples 
that are to be used for evaluation. The remaining point clouds are provided without 
annotations.

\subsection{Removing Occlusions}
Utilizing multiple sensors to capture the same scene inevitably leads to occlusions~\cite{9981654}.
To filter out occluded LiDAR points we follow~\cite{vpp4dc}. That method uses a sliding 
window, that moves on top of the depth points projected to the image plane. Points at a larger distance
than the nearest point contained in the sliding window by a given threshold are removed.
Considering the lower amount of LiDAR scan lines compared to~\cite{kitti-dataset}, we enlarge the 
sliding window to $10 \times 50$ pixels and set the distance threshold to $0.5$m. The points
marked for removal by the user are ignored in the process because removal of valid points
projected in the proximity of the dust clouds is not desired. The points to be removed by
the filtering method are marked as ``removed by algorithm" in the provided annotations.

\section{Experiments}
\label{sec:experiments}
We used the \emph{ShotcreteDepth} dataset ourselves to test various stereo, depth completion, and depth estimation baselines. The goal of these experiments was to elucidate the characteristics of the dataset.

\subsection{Experimental Setup}
For all evaluated methods, inference was performed at a resolution of $640\times480$ pixels.
The predictions were then upscaled and metrics computed at the original resolution of $1280\times960$.
The annotated and filtered LiDAR point clouds serve as the ground-truth for model evaluation.
The depth completion models were provided 500 uniformly sampled depth points $C$ originating 
from stereo matching. The same depth points were also used to align scale $a$ and shift $b$ 
of the affine-invariant predictions $D$ of the depth estimation models. The alignment was 
performed by minimizing the least square error:
\begin{equation}
(a, b) = \arg\min_{a,b}\sum_{i\in\Omega}\left(a\,D_{i}+b-C_{i}\right)^{2}
\end{equation}
\noindent The evaluation of stereo matching models requires ground-truth depth to be converted 
to disparity. Furthermore, disparity originating from the stereo matching needs to be converted 
to depth for sampling the sparse depth maps for the depth completion methods. This can be 
achieved by the formula:
\begin{equation}
    \it{depth} = \frac{baseline \times focal~length}{disparity}
\end{equation}
\noindent where the baseline is given by the stereo setup and focal length is determined 
in the camera calibration process.

\subsection{Metrics}
Following the standard practices, for depth estimation models we report absolute relative error
$\text{REL} = \frac{1}{N}\sum_i\lvert\frac{\mathbf{d}_i - \mathbf{g}_i}{\mathbf{g}_i}\rvert$
and $\delta_1$ percentage of points $i$ where 
$\max\left(\frac{\mathbf{d}_i}{\mathbf{g}_i},\frac{\mathbf{g}_i}{\mathbf{d}_i}\right) < 1.25$.
$N$ denotes count of pixels in an image, $\mathbf{d}_i$ are prediction pixels and 
$\mathbf{g}_i$ are ground-truth pixels. For depth completion models we report root 
mean squared error $\text{RMSE} = \sqrt{\frac{1}{N} \sum_{i} \lvert \mathbf{d}_i - \mathbf{g}_i \rvert^{2}}$ 
and mean absolute error $\text{MAE} = \frac{1}{N} \sum_{i} \lvert \mathbf{d}_i - \mathbf{g}_i \rvert$.
The values of both, RMSE and MAE, are always provided in meters. RMSE and MAE metrics in do not
fully capture the model's ability to produce good quality estimates, especially when the ground-truth
is sparse. Thus, we also assess boundary accuracy following~\cite{sharpdepth,ibims-dataset} 
and report Pseudo Depth Boundary Error (PDBE) accuracy $\mathcal{E}_{\text{PDBE}}^{\text{acc}}$ and 
completeness $\mathcal{E}_{\text{PDBE}}^{\text{comp}}$. We extract the ground-truth edges
for PDBE metrics from stereo matching. For stereo matching methods we report end-point 
error $\text{EPE} = \frac{1}{N} \sum_{i} \lvert \mathbf{d}_i - \mathbf{g}_i \rvert$
expressed in pixels. The formulaic definition of EPE matches MAE with the exception that the prediction 
and ground-truth are in disparity space. Additionally, D1 is percentage of elements
with error greater than 3 pixels and greater than $5\%$ of the ground-truth disparity. 
For all evaluated methods we provide runtime in seconds. Timing was performed
on Nvidia GeForce 4090. Model sizes are in millions of parameters.

\subsection{Models}
For FoundationStereo~\cite{foundation-stereo} we evaluated the ViT-Large checkpoint ``23-51-11''.
For StereoAnywhere~\cite{stereo-anywhere} we utilized the checkpoint pretrained on 
SceneFlow~\cite{sceneflow-dataset}. For RAFT-Stereo~\cite{raft-stereo} we downloaded the 
Middlebury checkpoint recommended for in-the-wild images. For Marigold-SSD~\cite{needforspeed}
we use the model trained on density range $\left[0.16\%, 5\%\right]$. Marigold-DC~\cite{marigold-dc} 
utilizes the Marigold~\cite{marigold} v1.0 checkpoint. For VPP4DC~\cite{vpp4dc} 
we used the model trained from scratch on SceneFlow~\cite{sceneflow-dataset}. 
For Depth Anything v3~\cite{depth-anything-v3}, we evaluated GIANT-1.1 
checkpoint. Marigold-E2E~\cite{marigold-e2e} weights were 
downloaded from HuggingFace. For MoGe-2~\cite{moge2} we
used ViT-Large checkpoint. The depth image
settings of the rc\_visard camera were set to defaults, with the exception of ``Quality" being
set to ``Full" and enabled ``Double-Shot" and ``Static".

\subsection{Stereo Matching}
\begin{figure}
    \centering
    \begin{subfigure}{.32\linewidth}
        \centering
        \includegraphics[width=1.0\linewidth]{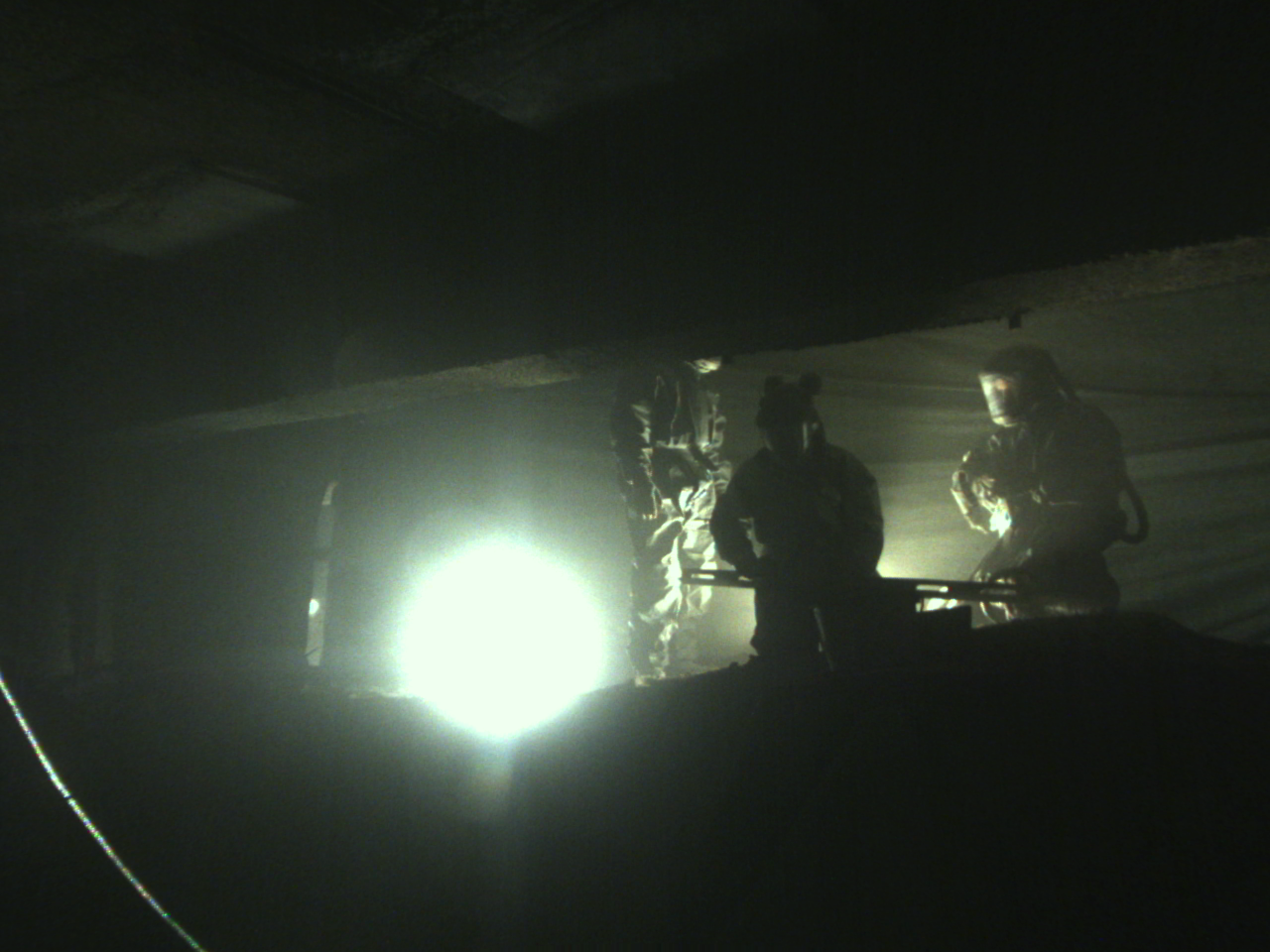}\\[1.2mm]
        \includegraphics[width=1.0\linewidth]{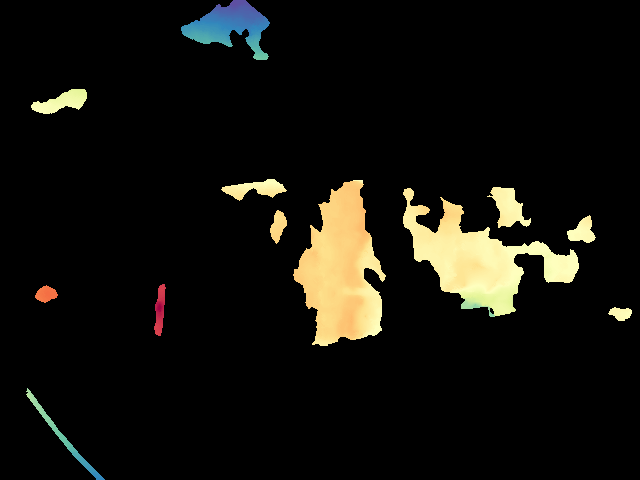}\\[1.2mm]
        \includegraphics[width=1.0\linewidth]{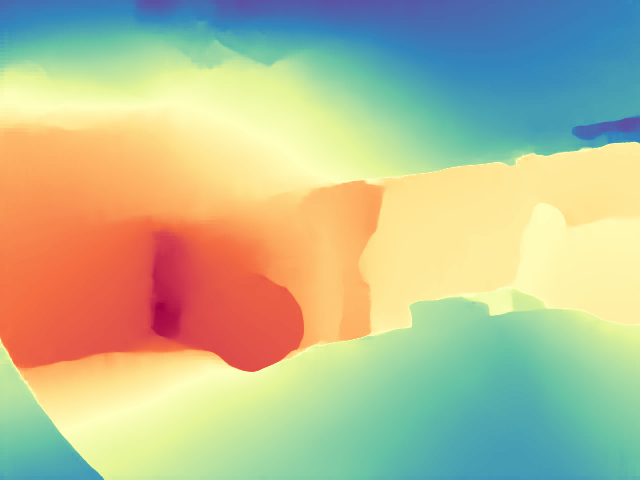}\\[1.2mm]
        \includegraphics[width=1.0\linewidth]{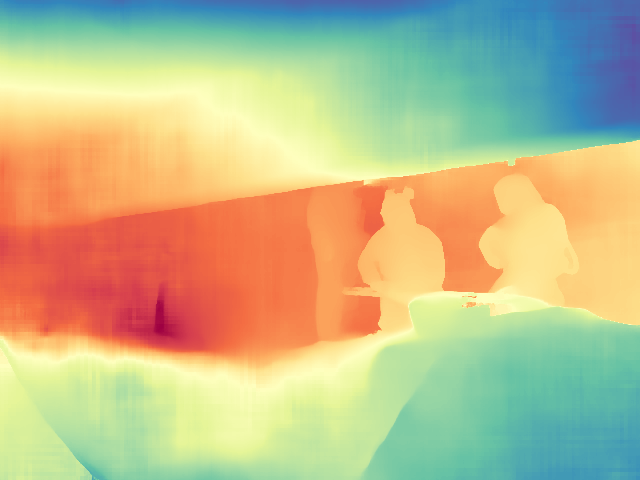}\\[1.2mm]
        \includegraphics[width=1.0\linewidth]{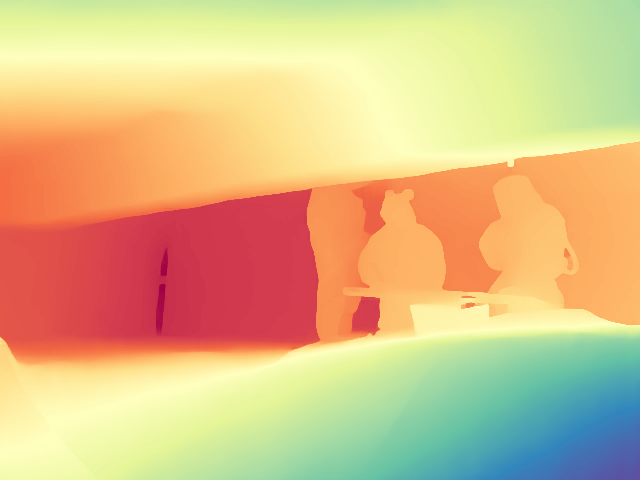}
    \end{subfigure}
    \begin{subfigure}{.32\linewidth}
        \centering
        \includegraphics[width=1.0\linewidth]{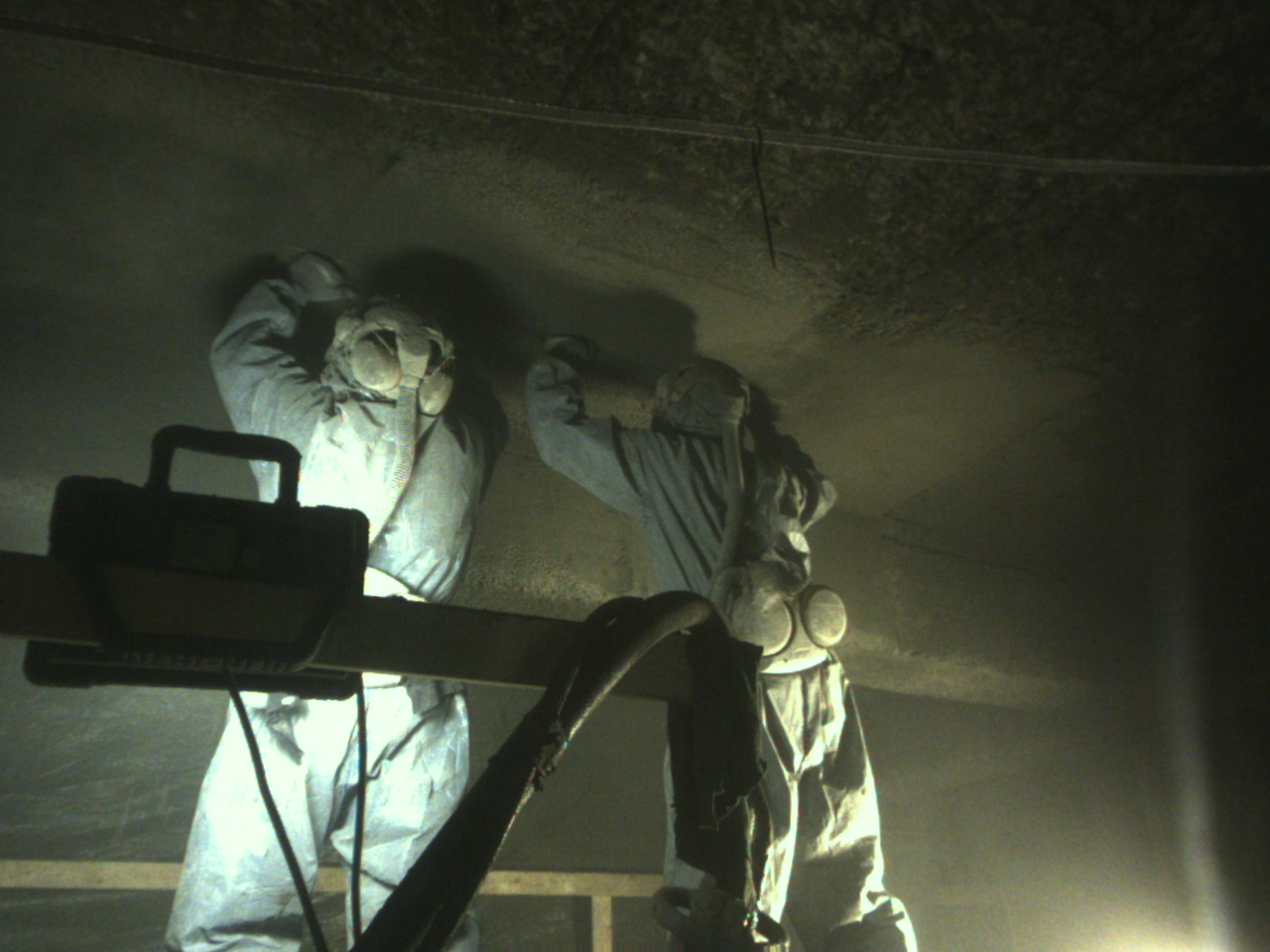}\\[1.2mm]
        \includegraphics[width=1.0\linewidth]{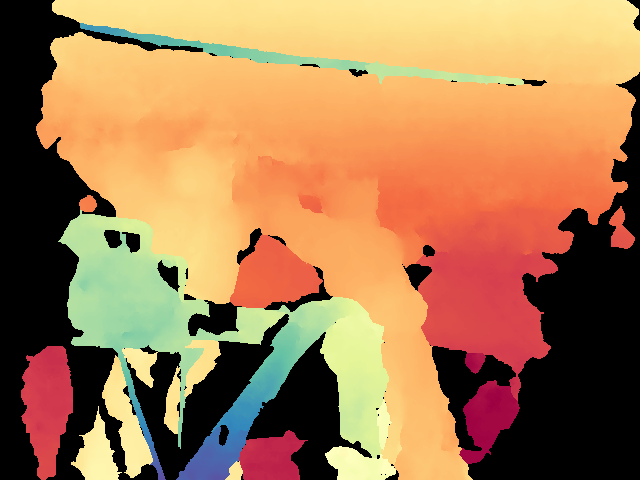}\\[1.2mm]
        \includegraphics[width=1.0\linewidth]{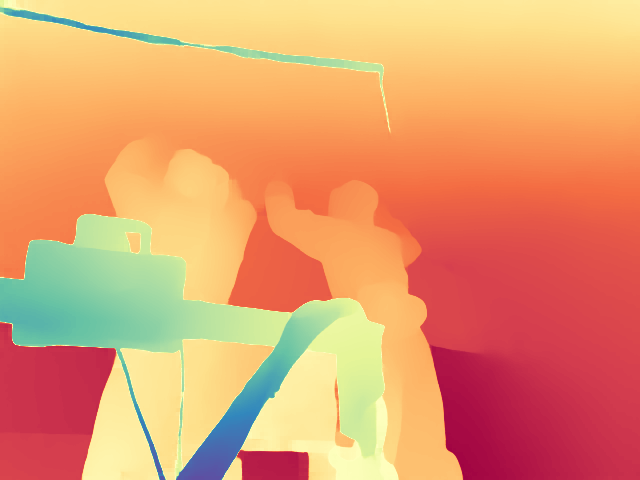}\\[1.2mm]
        \includegraphics[width=1.0\linewidth]{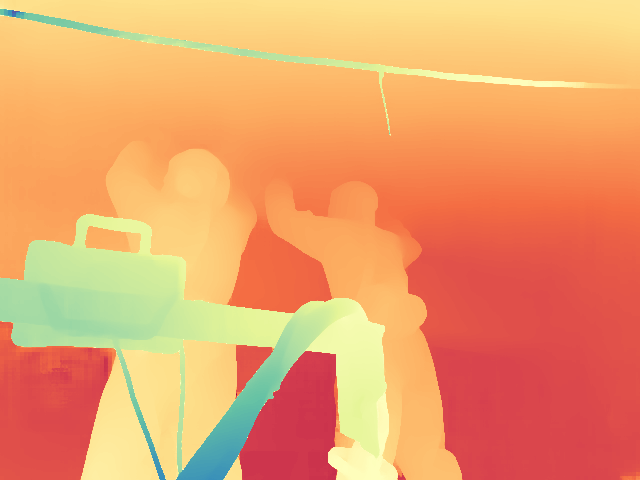}\\[1.2mm]
        \includegraphics[width=1.0\linewidth]{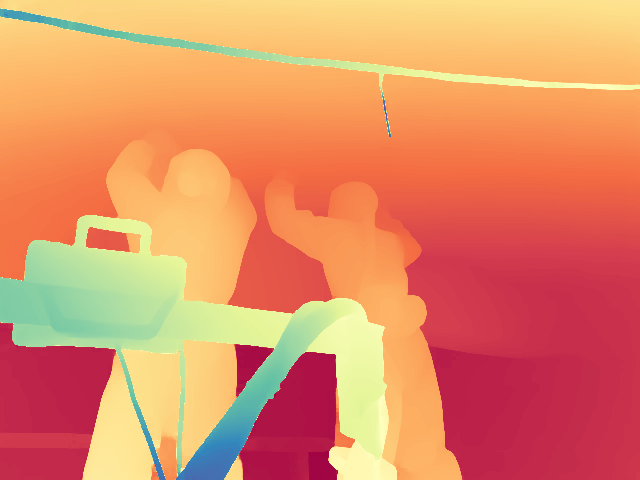}
    \end{subfigure}
    \begin{subfigure}{.32\linewidth}
        \centering
        \includegraphics[width=1.0\linewidth]{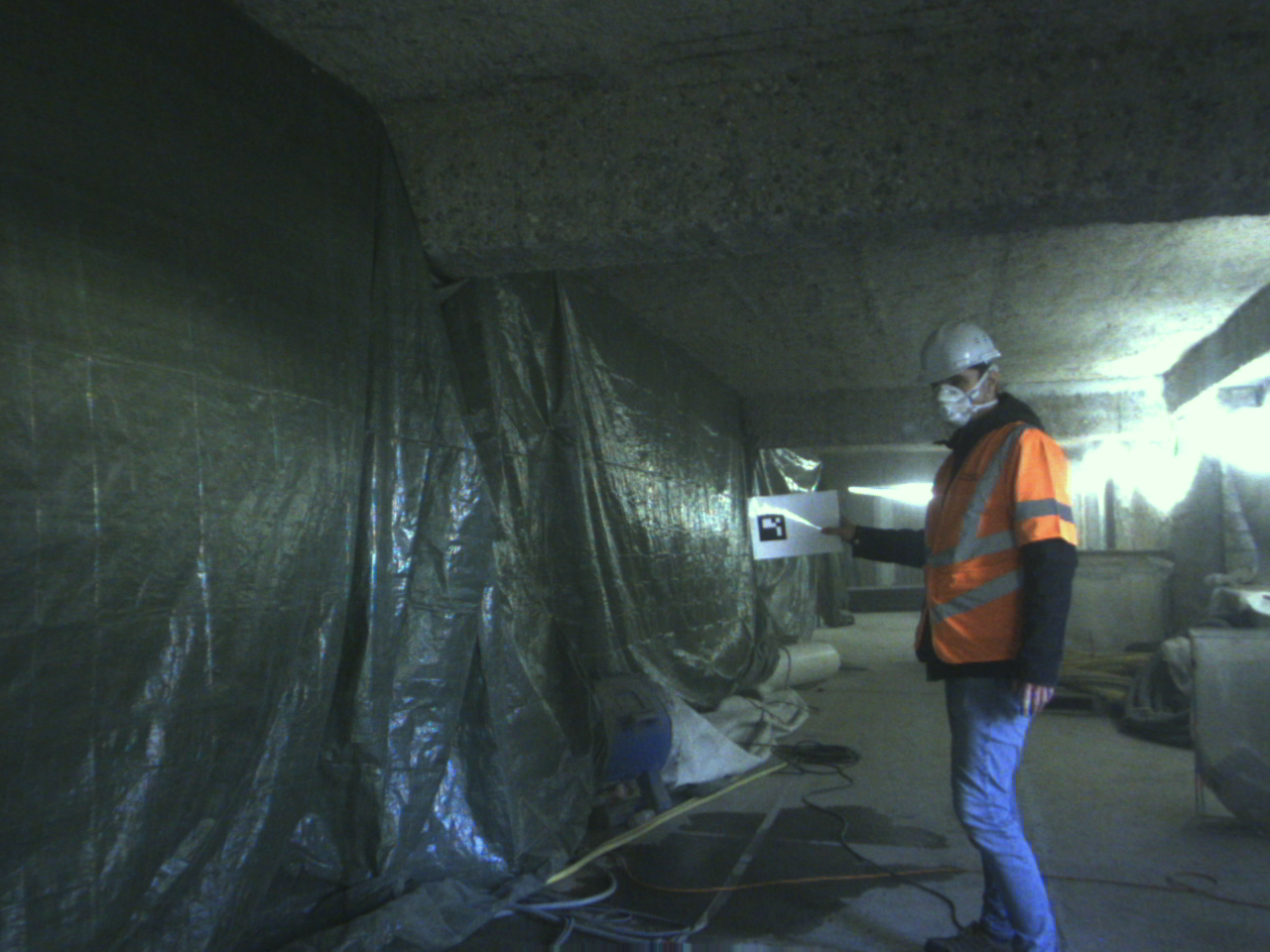}\\[1.2mm]
        \includegraphics[width=1.0\linewidth]{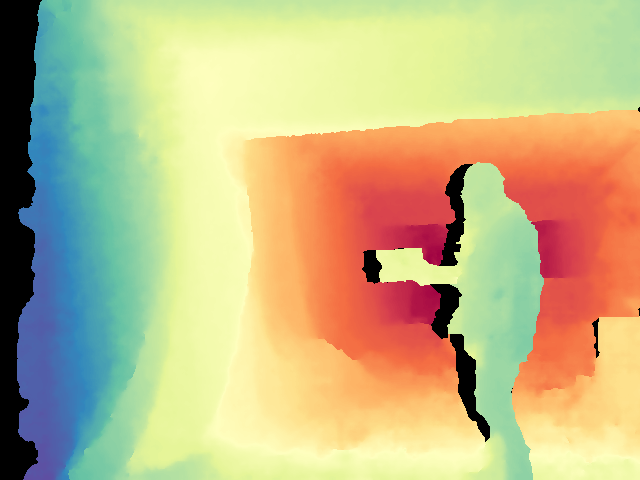}\\[1.2mm]
        \includegraphics[width=1.0\linewidth]{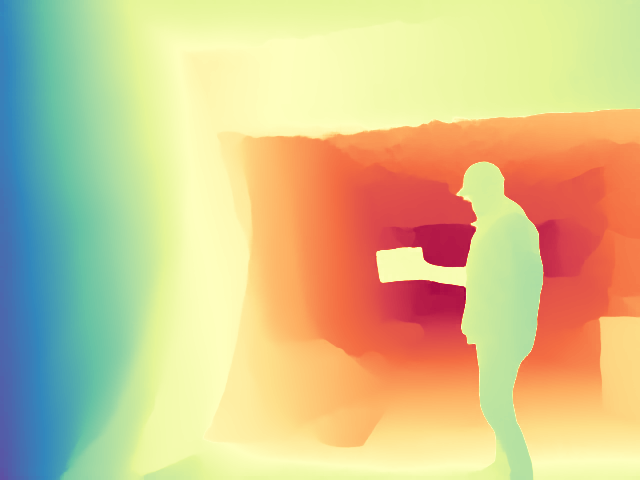}\\[1.2mm]
        \includegraphics[width=1.0\linewidth]{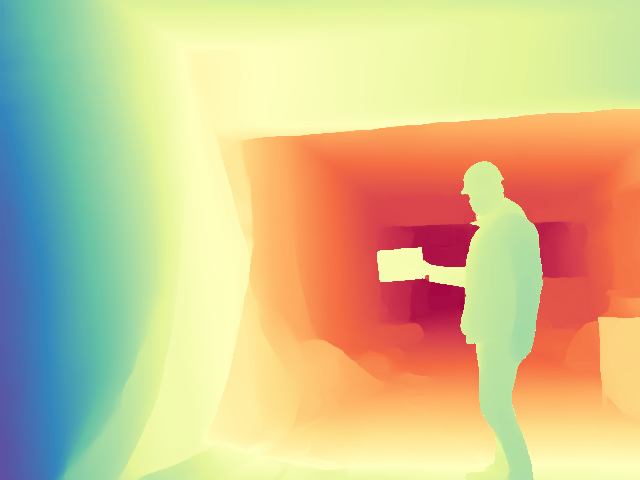}\\[1.2mm]
        \includegraphics[width=1.0\linewidth]{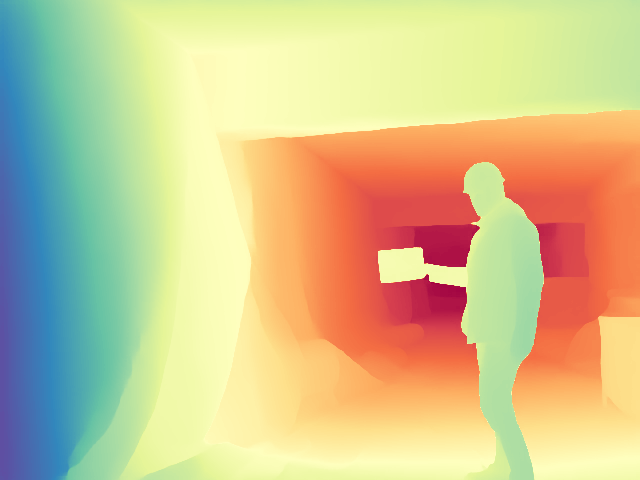}
    \end{subfigure}
	\caption{
        \textbf{Comparing stereo matching methods}. From top to bottom: left RGB image, 
        disparity maps computed by rc\_visard stereo matching, RAFT-Stereo~\cite{raft-stereo},
        FoundationStereo~\cite{foundation-stereo} and Stereo Anywhere~\cite{stereo-anywhere}.
    }
    \label{fig:qualitative-stereo-matching}
\end{figure}

The quantitative results evaluating the selected stereo matching methods are presented 
in Tab.~\ref{tab:stereo-matching}. In Fig.~\ref{fig:qualitative-stereo-matching} 
we compare the methods qualitatively. While stereo matching based on Semi-global 
Matching~\cite{sgm} is optimized enough to run on embedded hardware, it often produces
incomplete depth maps. The more computationally heavy-weight neural networks 
are more robust in the challenging environmental conditions captured in our
dataset. We can observe, that the methods based on neural networks can deal
with occlusions and provide more visually distinct edges on the object boundaries, even in extremely dark (cf. Fig.~\ref{fig:qualitative-stereo-matching}-left) or overexposed  (cf. Fig.~\ref{fig:qualitative-stereo-matching}-middle) parts of images.
\begin{table}[b]
\caption{\textbf{Evaluation of stereo matching methods}: three neural-network based 
approaches (RAFT - RAFT-Stereo~\cite{raft-stereo}, FS - FoundationStereo~\cite{foundation-stereo}, 
SA - Stereo Anywhere~\cite{stereo-anywhere}) and the proprietary SGM~\cite{sgm} implementation
running on the Roboception rc\_visard 160 camera. Timing was performed on Nvidia GeForce 4090, 
while runtime of the camera's stereo matching given by the FPS of the camera.
}
\label{tab:stereo-matching}
\centering
\begin{tabular}{l|cccc}
\toprule
\multirow{2}{*}{Metrics} & \multicolumn{4}{c}{Methods} \\
                 & RAFT  & FS    & SA    & rc\_visard \\
\midrule
EPE$\downarrow$  & 2.449 & 2.439 & 2.328 & 2.276      \\
D1$\downarrow$   & 0.130 & 0.129 & 0.103 & 0.106      \\
MAE$\downarrow$  & 0.337 & 0.327 & 0.311 & 1.467      \\
RMSE$\downarrow$ & 0.729 & 0.741 & 0.795 & 2.207      \\
\midrule
Coverage         & 100\% & 100\% & 100\% & 61\%       \\
\midrule
Parameters       & 11 M  & 375 M & 347 M & -          \\
Runtime          & 0.141 & 0.156 & 0.307 & 0.333      \\
\bottomrule
\end{tabular}
\end{table}

\subsection{Depth Completion}
\begin{table}
\caption{\textbf{Evaluation of three depth completion methods}: Marigold-DC~\cite{marigold-dc} 
for a single run ($1^{\text{st}}$ value) and ensemble of 10 runs ($2^{\text{nd}}$ value),
Marigold-SSD~\cite{needforspeed} and VPP4DC~\cite{vpp4dc}. 
}
\label{tab:depth-completion}
\centering
\begin{tabular}{l|ccc}
\toprule
\multirow{2}{*}{Metrics}                            & \multicolumn{3}{c}{Methods}              \\
                                                    & Marigold-DC      & Marigold-SSD & VPP4DC \\
\midrule
MAE$\downarrow$                                     & 0.364 / 0.325    & 0.422        & 0.360  \\[0.5ex]
RMSE$\downarrow$                                    & 0.893 / 0.793    & 0.868        & 0.790  \\[0.5ex]
$\mathcal{E}_{\text{PDBE}}^{\text{acc}}\downarrow$  & 1.636 / 1.189    & 1.576        & 2.076  \\[0.5ex]
$\mathcal{E}_{\text{PDBE}}^{\text{comp}}\downarrow$ & 4.687 / 5.780    & 21.107       & 10.783 \\[0.5ex]
\midrule
Parameters                                          & 950 M            & 989 M        & 11 M   \\[0.5ex]
Runtime                                             & 24.479 / 254.284 & 0.382        & 0.116  \\
\bottomrule
\end{tabular}
\end{table}

\begin{figure*}[t]
    \centering
    \begin{subfigure}{.195\linewidth}
        \centering
        \includegraphics[width=1.0\linewidth]{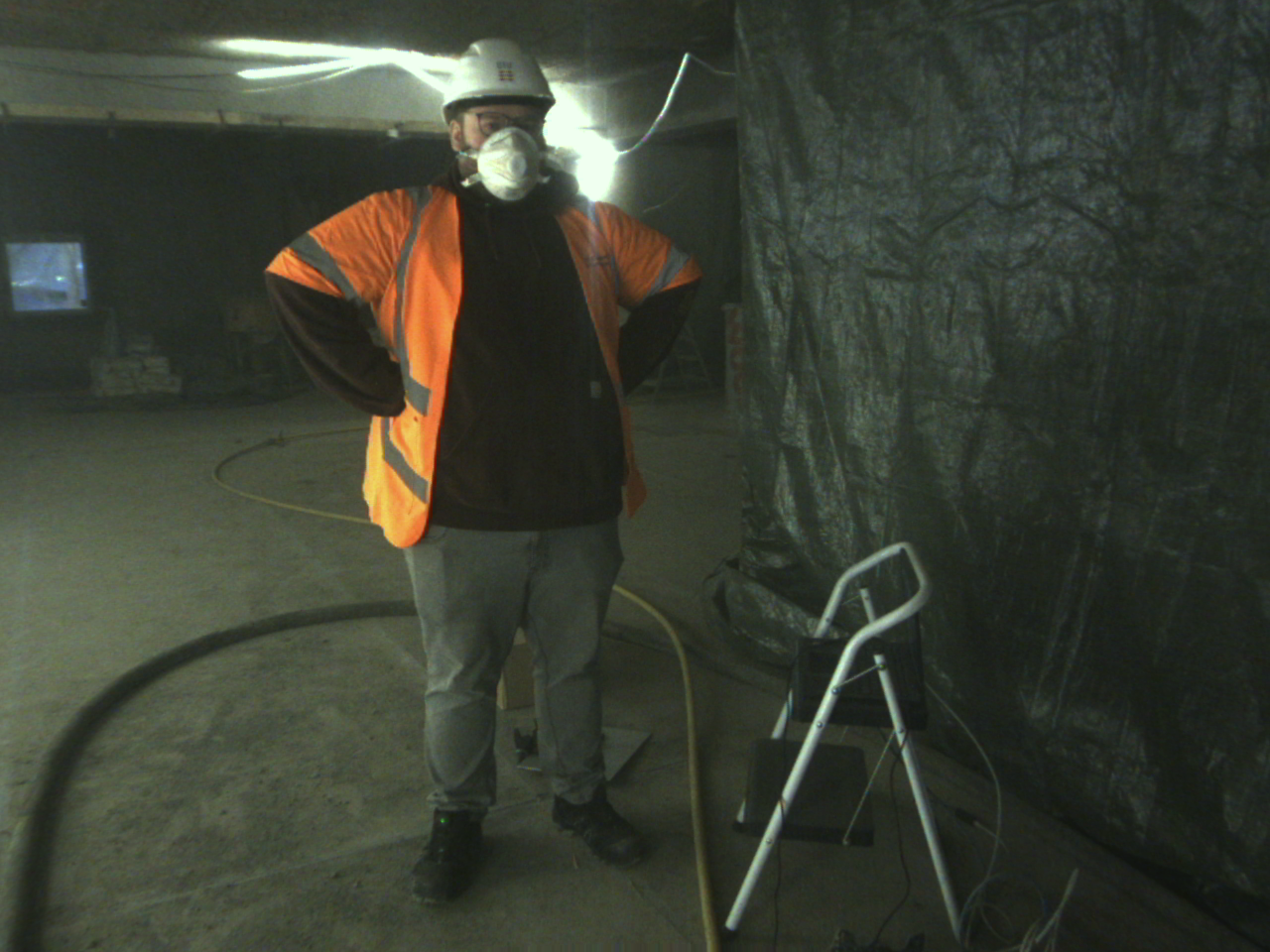}\\[1.2mm]
        \includegraphics[width=1.0\linewidth]{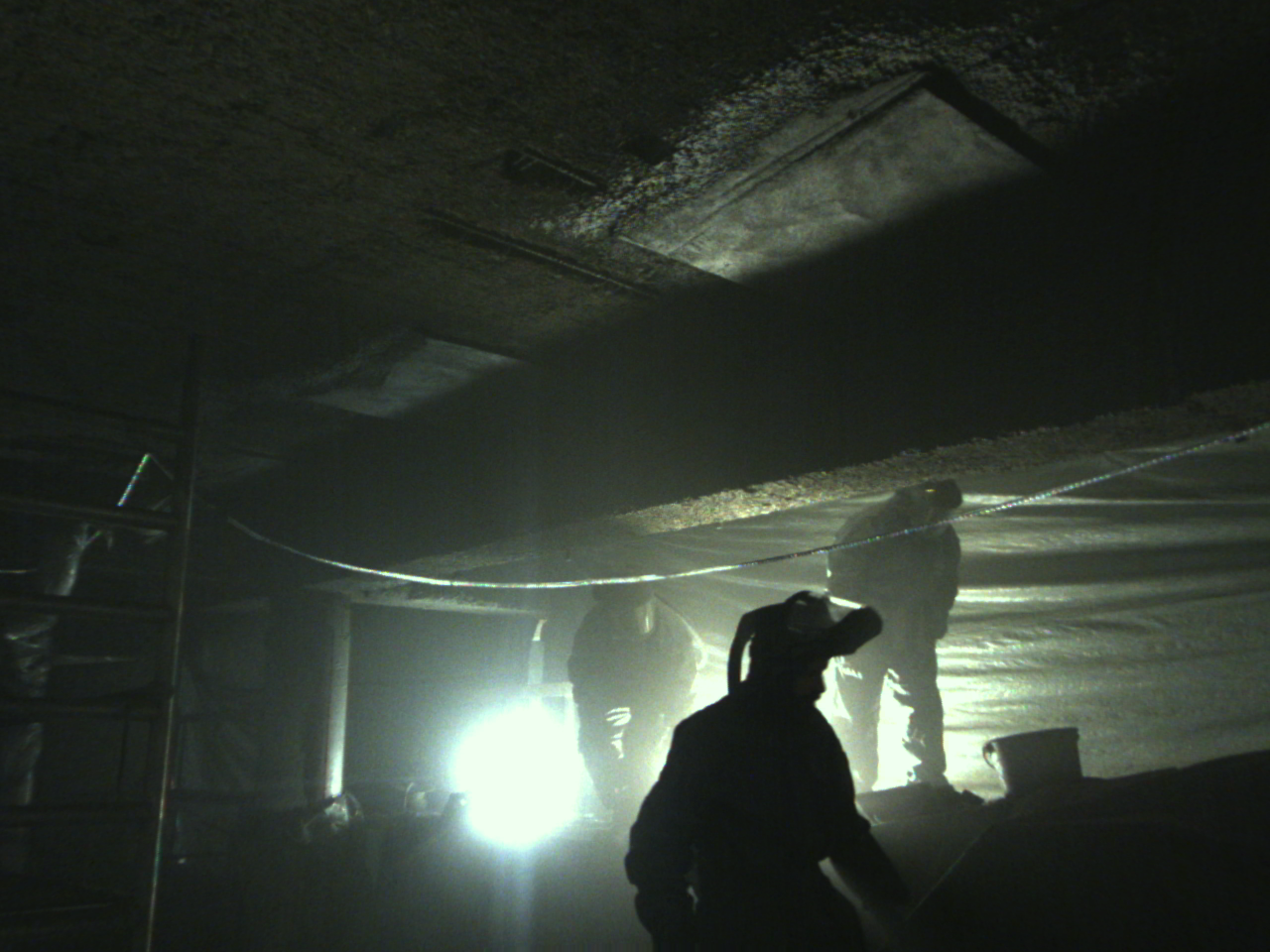}
        \caption{RGB}
    \end{subfigure}
    \begin{subfigure}{.195\linewidth}
        \centering
        \includegraphics[width=1.0\linewidth]{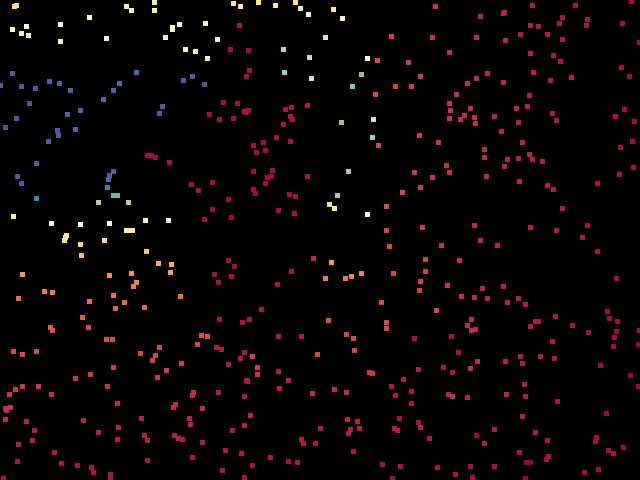}\\[1.2mm]
        \includegraphics[width=1.0\linewidth]{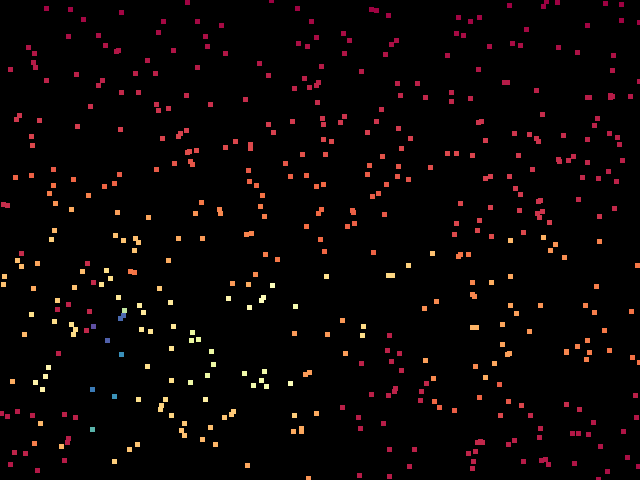}
        \caption{Sparse Depth}
    \end{subfigure}
    \begin{subfigure}{.195\linewidth}
        \centering
        \includegraphics[width=1.0\linewidth]{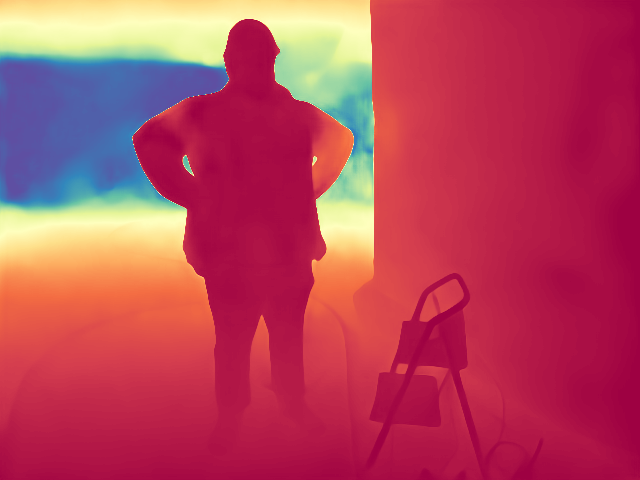}\\[1.2mm]
        \includegraphics[width=1.0\linewidth]{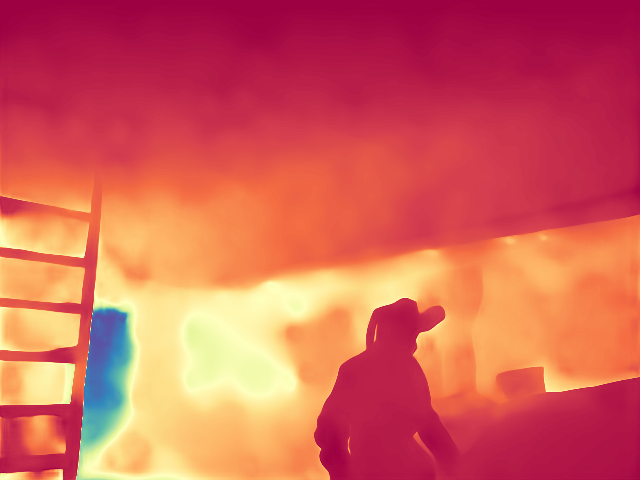}
        \caption{Marigold-SSD}
    \end{subfigure}
    \begin{subfigure}{.195\linewidth}
        \centering
        \includegraphics[width=1.0\linewidth]{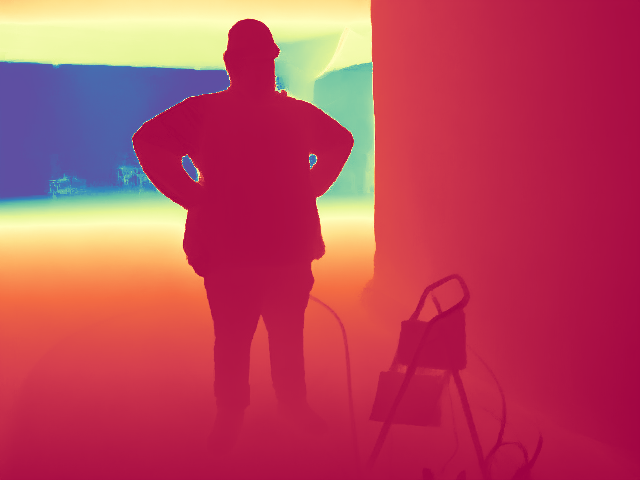}\\[1.2mm]
        \includegraphics[width=1.0\linewidth]{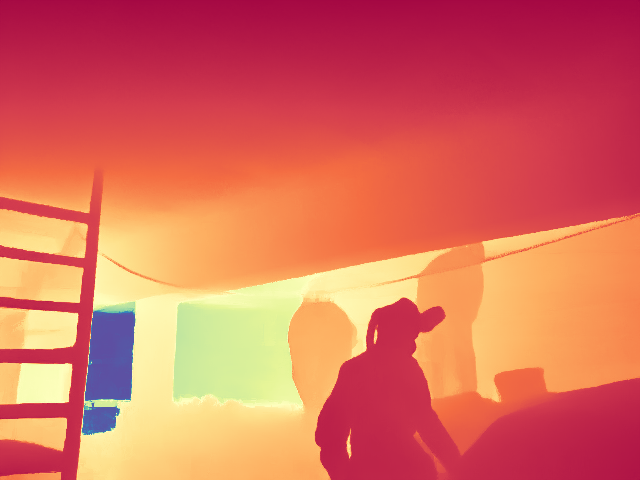}
        \caption{Marigold-DC}
    \end{subfigure}
    \begin{subfigure}{.195\linewidth}
        \centering
        \includegraphics[width=1.0\linewidth]{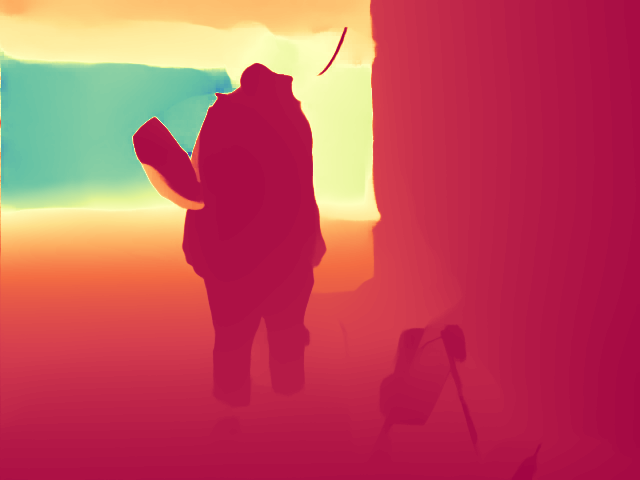}\\[1.2mm]
        \includegraphics[width=1.0\linewidth]{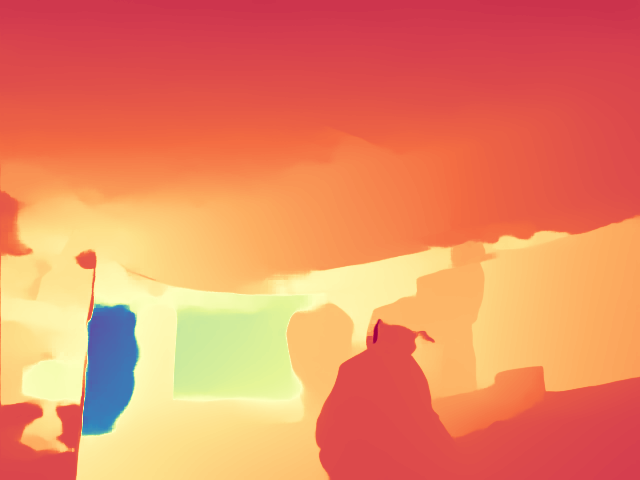}
        \caption{VPP4DC}
    \end{subfigure}
	\caption{
        \textbf{Qualitative results for depth completion methods.}
    }
    \label{fig:qualitative-depth-completion}
\end{figure*}

The results for depth completion are presented in Tab.~\ref{tab:depth-completion}.
The depth methods were used to complete depth maps sampled from stereo matching. 
The Stereo Anywhere method was selected as the source of depth, because it achieves
the lowest EPE and D1, which are the common metrics for evaluation of the matching 
algorithms. 500 depth points were sampled uniformly. Stereo Anywhere was also used
as the source of ground-truth edges for PDBE metrics, because it produces visually
cleanest output (see Fig.~\ref{fig:qualitative-stereo-matching}). The qualitative 
results including visualization of the sampled sparse depth are presented in 
Fig.~\ref{fig:qualitative-depth-completion}.

\subsection{Depth Estimation}
\begin{table}
\caption{\textbf{Evaluation of depth estimation methods}: 
DepthAnything v3~\cite{depth-anything-v3}, Marigold-E2E~\cite{marigold-e2e},
MoGe-2~\cite{moge2} with scale and shift alignment ($1^{\text{st}}$ value) and 
without ($2^{\text{nd}}$ value).
}
\label{tab:depth-estimation}
\centering
\begin{tabular}{l|ccc}
\toprule
\multirow{2}{*}{Metrics} & 
\multicolumn{3}{c}{Methods} \\
                   & DepthAnything v3 & Marigold-E2E & MoGe-2        \\
\midrule
REL$\downarrow$    & 0.133            & 0.174        & 0.142 / 0.210 \\
$\delta_1\uparrow$ & 0.834            & 0.727        & 0.816 / 0.594 \\
MAE$\downarrow$    & 0.554            & 0.715        & 0.576 / 0.832 \\
RMSE$\downarrow$   & 0.852            & 1.016        & 0.879 / 1.134 \\
\midrule
Parameters         & 1356 M           & 950 M        & 326 M         \\
Runtime            & 0.095            & 0.175        & 0.057         \\
\bottomrule
\end{tabular}
\end{table}

\begin{figure*}
    \centering
    \begin{subfigure}{.245\linewidth}
        \centering
        \includegraphics[width=1.0\linewidth]{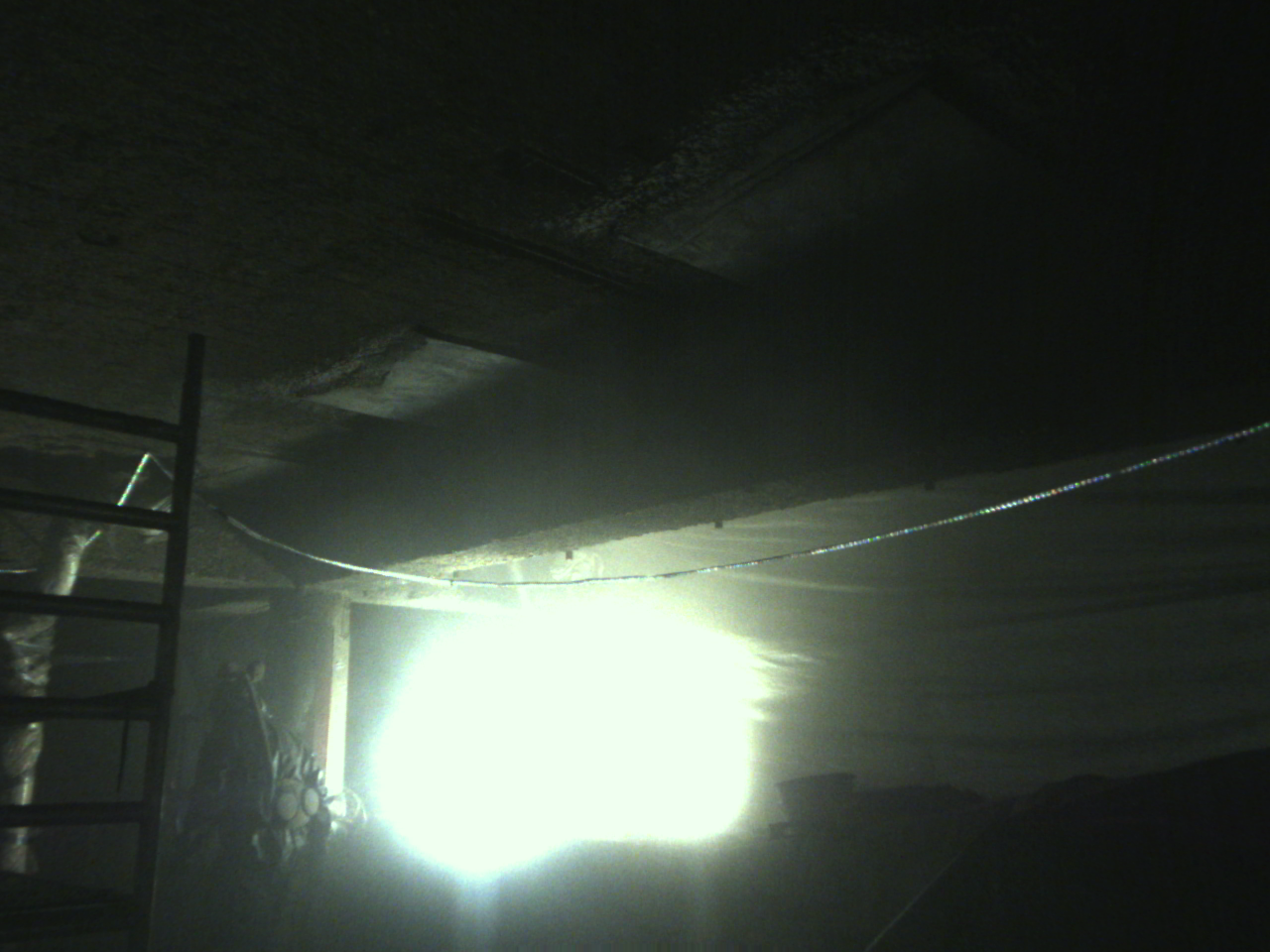}\\[1.2mm]
        \includegraphics[width=1.0\linewidth]{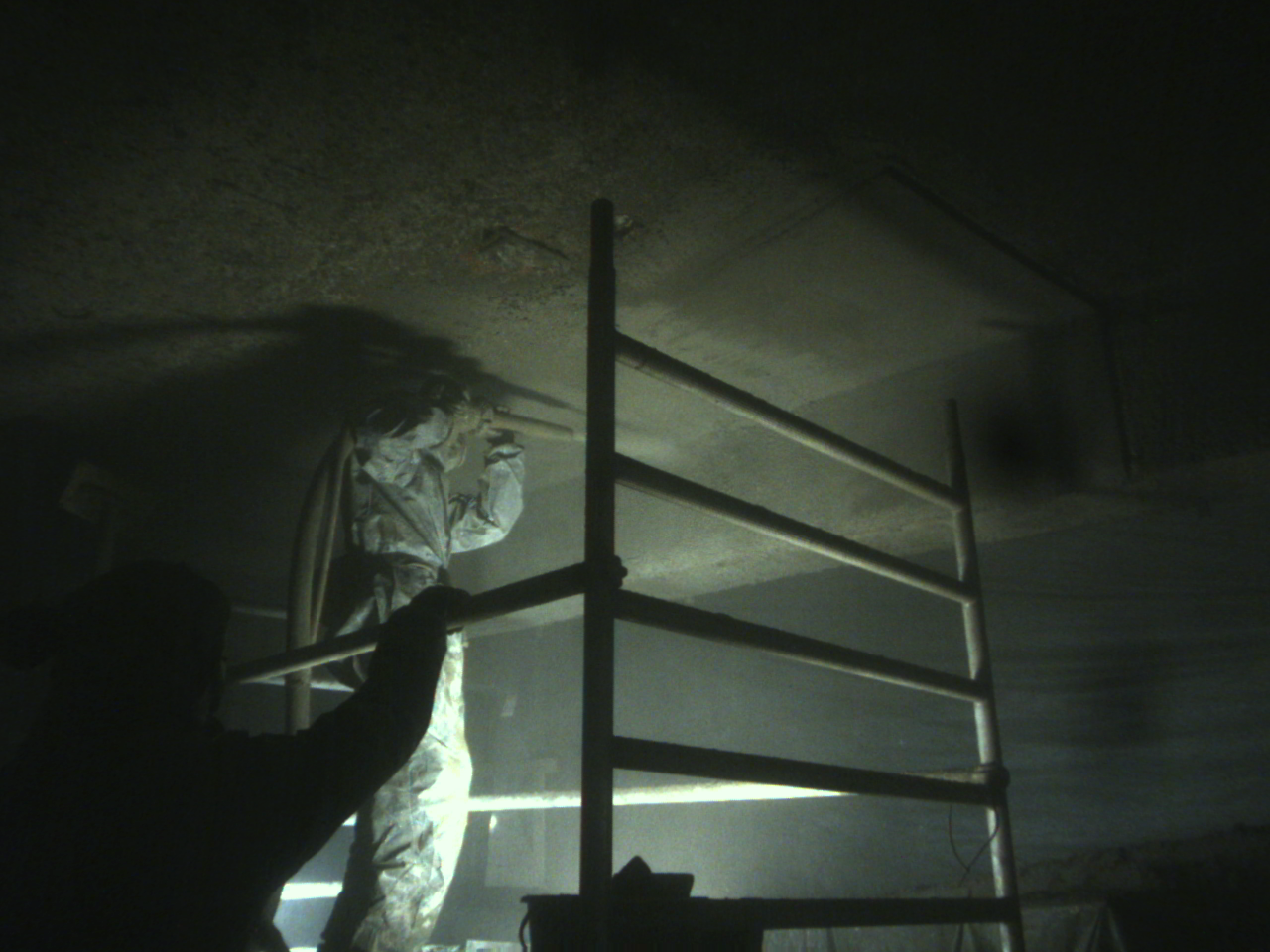}
        \caption{RGB}
    \end{subfigure}
    \begin{subfigure}{.245\linewidth}
        \centering
        \includegraphics[width=1.0\linewidth]{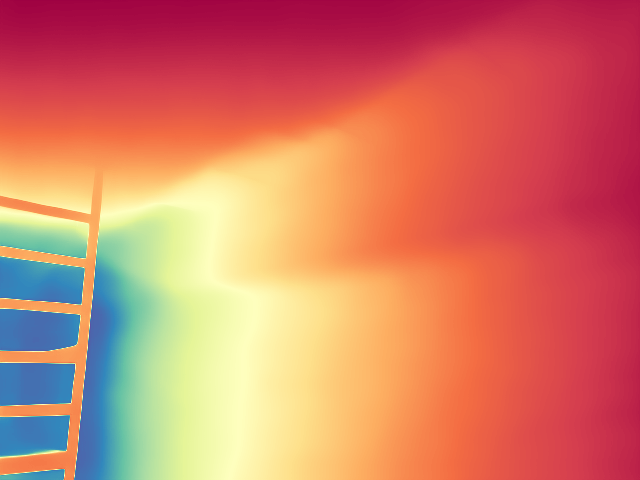}\\[1.2mm]
        \includegraphics[width=1.0\linewidth]{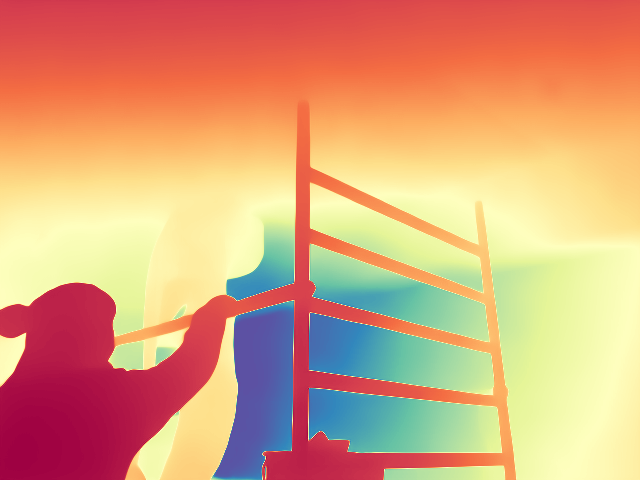}
        \caption{Marigold-E2E}
    \end{subfigure}
    \begin{subfigure}{.245\linewidth}
        \centering
        \includegraphics[width=1.0\linewidth]{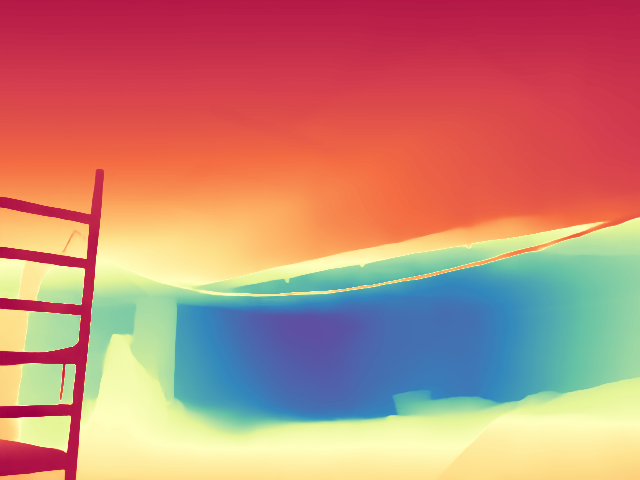}\\[1.2mm]
        \includegraphics[width=1.0\linewidth]{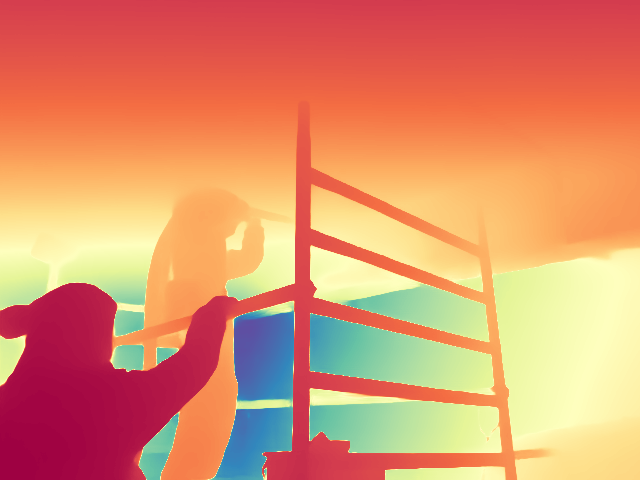}
        \caption{Depth Anything v3}
    \end{subfigure}
    \begin{subfigure}{.245\linewidth}
        \centering
        \includegraphics[width=1.0\linewidth]{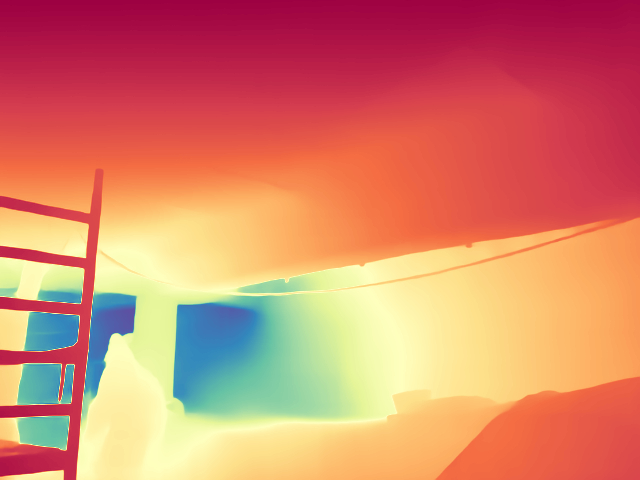}\\[1.2mm]
        \includegraphics[width=1.0\linewidth]{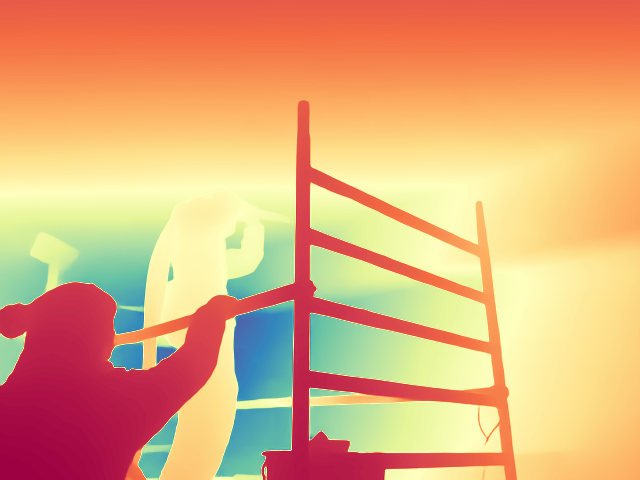}
        \caption{MoGe-2}
    \end{subfigure}
	\caption{
        \textbf{Qualitative results for depth estimation methods.}
    }
    \label{fig:qualitative-depth-estimation}
\end{figure*}

We evaluated three depth estimation models; the results are summarized in 
Tab.~\ref{tab:depth-estimation}. The sparse depth maps sampled for the 
depth completion methods in the previous subsection were used to align
shift and scale of the affine invariant estimates of Depth Anything 
v3~\cite{depth-anything3} and Marigold-E2E~\cite{marigold-e2e}.
Since MoGe-2~\cite{moge2} predicts metric depth we evaluated it with
and without shift and scale adjustment. The qualitative results are
presented in Fig.~\ref{fig:qualitative-depth-estimation}.

\section{Discussion and Conclusion}
Our experiments in Sec.~\ref{sec:experiments}, have shown that depth perception is indeed possible in the challenging conditions of shotcreting environments. Each one of the tested approaches---3 stereo matching approaches, 3 depth completion methods and 3 depth estimation methods---has exhibited merits in the corresponding task. While computational efficiency and near-real-time operation is feasible, this comes at the expense of output accuracy. Contrary, large deep learning-based models are characterized by inferior runtimes (even when executed on dedicated GPU-enabled workstations), but their accuracy is typically growing with their size for all 3 considered tasks.

In this work, we have introduced \emph{ShotcreteDepth}, a bi-modal dataset for evaluation and development of depth 
perception methods and used it to test 9 state-of-the-art deep learning 
methods. Our dataset captures the niche shotcreting environment;
an environment characterized by high turbidity. Additionally, we developed a lightweight
annotation tool for 3D point clouds, which used to remove the dust clouds
observed by LiDAR from the evaluation. Stereo cameras and LiDARs are 
fundamentally distinct and better suited for different scenarios.
We see a potential to achieve the best depth perception by their fusion,
which would require an additional source of depth for evaluation.

\bibliographystyle{ieee_fullname}
\bibliography{main}

\end{document}